\documentclass[conference]{IEEEtran}
\IEEEoverridecommandlockouts
\usepackage{float}
\usepackage{enumitem}

\usepackage[super]{nth}
\usepackage{amsmath,amssymb,amsfonts}
\usepackage{algorithmic}
\usepackage{graphicx}
\usepackage{textcomp}
\usepackage{xcolor}
\usepackage{hyperref}
\usepackage{biblatex}
\def\BibTeX{{\rm B\kern-.05em{\sc i\kern-.025em b}\kern-.08em
    T\kern-.1667em\lower.7ex\hbox{E}\kern-.125emX}}
\begin{document}

\title{Context-aware 6D Pose Estimation of Known Objects using RGB-D data\\

\thanks{}
}
\makeatletter
\newcommand{\linebreakand}{%
  \end{@IEEEauthorhalign}
  \hfill\mbox{}\par
  \mbox{}\hfill\begin{@IEEEauthorhalign}
}
\makeatother

\author{
\IEEEauthorblockN{Ankit Kumar, Priya Shukla, Vandana Kushwaha and G.C. Nandi}
\IEEEauthorblockA{\textit{Center of Intelligent Robotics, } \\
Indian Institute of Information Technology Allahabad, Prayagraj, India \\
ramuk.tikna10@gmail.com, priyashuklalko@gmail.com, kush.vandu@gmail.com and gcnandi@iiita.ac.in }


}

\maketitle

\begin{abstract}
6D object  pose  estimation  has been  a  research  topic  in  the  field  of  computer vision and robotics.  Many modern world applications like robot grasping, manipulation,  autonomous navigation etc, require the correct pose of objects present in a scene to perform their specific task.  It becomes even harder when the objects are placed in a cluttered scene and the level of occlusion is high.  Prior works have tried to overcome this problem but could not achieve accuracy that can be considered reliable in  real-world applications. 
In this paper, we present an architecture that, unlike prior work, is context-aware. 
It utilizes the context information available to us about the objects. 
Our proposed architecture treats the objects separately according to their types i.e; symmetric and non-symmetric.
A deeper estimator and refiner network pair is used for non-symmetric objects as compared to symmetric due to their intrinsic differences.  Our experiments show an enhancement in the accuracy of about 3.2 \% over the LineMOD dataset, which is considered a benchmark for pose estimation in the occluded and cluttered scenes, against the prior state-of-the-art DenseFusion. Our results also show that the inference time we got is sufficient for real-time usage.
\end{abstract}


\section{\textbf{Introduction}}
Pose estimation of objects with 6 Degree of Freedom (6DoF) is among the most important initial steps for many modern real-world applications like robotic hand grasping and manipulation of objects \cite{3,4}, autonomous navigation \cite{6,7}, and the exciting domain of augmented reality \cite{18,19}. We see that robotic hands are being used in the automation of manufacturing industries, but these robots are not intelligent enough as they get their input in a strictly defined manner, which means the coordinates and orientation is kept fixed and even for a slight change in the orientation or the shape and size of the input, the whole setup may need to be redefined. In the domain of autonomous navigation, we need proper lighting on the obstructions for them to be recognized and dealt with. In the world of augmented reality, the occluded area of the components in the scene also plays a very important role we can't simply ignore it as we need to map the augmented imagery with the real world based on its coordinates and orientation.\par

From the above use cases, we can infer that we need a solution that can deal with the objects that are of varying shape and size along with the different types of surface properties i.e; textures. Also, the solution should be able to deal with the objects having heavy occlusion, varying light exposure, imperfect lighting and   sensor noise. All this needs to be done in real-time in order to cope up with the needs of the real-world applications requirement.\par

With the advancement in sensor building technology we now have cheap sensors that can capture RGB-D images efficiently and in real-time, so we can now device better approaches in solving our problems using RGB-D data as opposed to when we only use RGB data.\par

Initially, the classical approaches were to somehow extract the features from the RGB-D data, perform the corresponding grouping and, then using it for hypothesis verification \cite{11,12,13,14,15}. But these methods were used to rely on handcrafted features and a fixed matching procedure which resulted in limited performances in presence of light variation and occlusion. Then after success in visual recognition gave rise to data-driven methods that used deep multi-layered perceptron networks for pose estimation of known objects from RGB-D data, some examples are PoseCNN \cite{16} and MCN \cite{17}.\par

But the major shortcoming of all these methods was that they required time-consuming post-processing refinement steps to completely use the essence of 3D information, such as ICP (Iterative Closest Point)\cite{18} in PoseCNN and a multilevel-view hypothesis verification in MCN. All these refinement steps were not possible to be made efficient and real-time performing. Moreover, in the domain of autonomous driving, some promising technologies that have proved their efficacy are Frustum PointNet \cite{19} and PointFusion \cite{7}. These models have shown good performance and in real-time too but these methods were found not much effective in heavy occlusion which is normal in the manipulation domains.\par

After these methods, a more advanced architecture was proposed named DenseFusion \cite{1}, where an end-to-end deep learning based architecture is used for the practically efficient solution for estimating the 6-DoF pose of known objects from RGB-D inputs. Before this, image crops to compute global features \cite{7}, or 2D boxes that bound the objects \cite{19} were used, but in the DenseFusion the core approach was to somehow take the essence of both RGB and Depth information and fuse their information and embed both to obtain the collective information at the per-pixel level. This fusion helped to reason about local appearance and orientation which in turn makes it able to handle heavy occlusions as well. And after the fusion, there is an integrated end-to-end learning framework that replaced the all expensive post-hoc refinement in terms of time. This is an iterative method that performs pose refinement.\par

In this work, we propose an enhanced model that is based on the core idea of DenseFusion, which identifies the context information present in the dataset which was unused in the previous methods and use it to obtain much better accuracy. The symmetric and non-symmetric objects are treated separately so that one does not alter the information gain from the other. We propose a separate pose estimator and refiner architecture that has shown improved performance in terms of accuracy keeping the inference time in the same order. 
We have improved the accuracy by  3.2 \% over the DenseFusion model for the LineMOD dataset\cite{20} which is one of the most popular benchmarks for 6D pose estimation. 
Our improved results signify the higher performance in highly cluttered scenes, scenes with variable lighting, and for occluded objects.\par

In summary, the contribution of our work is to improve the accuracy for 6D pose estimation of known objects using RGB-D data and its context information. The core architecture of DenseFusion\cite{1} is used as the starting point and then enhanced significantly to achieve higher accuracy without hindering the real-time speed being offered.

\section{\textbf{Related Work\\}}  

The field of Pose estimation is relatively new as the advent of RGB-D sensors and the Machine Learning based techniques are new which have shown promise in this domain. Upon researching one can find that there have been several efforts made in the past to correctly predict the pose of an object. Many have tried only RGB data, many only depth information and some have tried using a combination of both.\par

Initial methods that took shape using RGB data only, used to rely on key-points detection and matching with known object models \cite{21,22,4}. After some progress we started seeing methods that tried to solve the challenge in hand by learning to predict the 2D keypoints \cite{23,24,25,3} and solve the pose by PnP that uses RANSAC (RANdom SAmple Consensus), but it could not cope with the speed demanding tasks that needs to be performed in real time when the input is low textured or low resolution. Other methods that uses only RGB image to predict pose is by using CNN-based architectures \cite{27}. Some of the methods are Xiang et al. \cite{29, 30} which clusters 3D features from object models and then learns to predict pose according to the viewpoints. Mousavian et al. \cite{31} recovers poses by single view geometric constraints and predicts 3D object parameters. Sundermeyer et al. \cite{32} creates a codebook which contains encoding of orientations in a latent space then finds the best match from the codebook as the prediction. Despite putting in this much effort it was concluded that something extra along with RGB image is required in order to estimate object poses in even 3D let alone 6D.\par

Various efforts have been done over the time where the problem in hand i.e; pose estimation or 3D object detection is tried to be solved using depth information only which we also refer to as point clouds. As an example Song et al. \cite{33,34} estimates the poses of objects by featuring inputs with 3D ConvNets to generate 3D bounding box. It has been observed that these methods can pretty much accurately encode geometric information , but as a trade-off they are implicitly expensive, \cite{34} takes around 20 seconds which makes them unfit for real time applications.\par

Recently deep learning architectures have evolved very much such that it has enabled methods that can directly predict the poses of objects on point clouds or 3D data. For example Frustrum PointNet \cite{19} and VoxelNet \cite{35} uses PointNet \cite{36} like structure and have proven their performances as state-of-the-art on KITTI benchmark \cite{6}. DenseFusion \cite{1} also uses similar kind of architecture but unlike urban driving applications, datasets like YCB- video dataset also require keen observation on appearance along with geometric information, which is implemented in DenseFusion by 2D-3D sensor fusion architecture. The DenseFusion architecture is further improved in our work by segregating separate prediction networks for symmetric and non symmetric objects keeping the point cloud and colour embedding part unchanged. \par

Classical approaches use the image processing principles in which 3D features are extracted from the RGB-D input data then they perform hypothesis verification after doing correspondence grouping \cite{20,11,12,13,14,15}. However, these features are chosen based on the bag of words principle that are hardcoded beforehand or are learned by optimizing surrogate objectives \cite{14,15}. Certain newer methods took shape that directly estimates 6D poses from image data such as PoseCNN \cite{16} and some try to utilize the depth information by fussing the depth value as an additional channel such as Li et al. \cite{17}. But all such methods performance further confirms that depth information also needs to be handled carefully and not just appended with image data. However, these methods need expensive post-processing steps to rely on to reach their full capacity. The latest addition to this domain is DenseFusion which doesn't require such post-processing steps but also uses the RGB data along with depth. It uses a novel fusion mechanism to map colour and depth information which yielded much better results but still falls just short to provide accuracy that could be termed reliable for real world applications. Then our work which keeping uncompromised the speed of the DenseFusion and further improves the performance in terms of accuracy on the LineMOD dataset. 

\section{\textbf{Problem Statement}}
Our main focus here is on the real-world problem of estimating the 6D pose of the objects whose shape properties are known to us beforehand using the information collected as the colour image as well as the depth information. Lets now break down the problem statement to discuss it in details.\par

\begin{figure}[!ht]
    \center
    \includegraphics[scale=0.3]{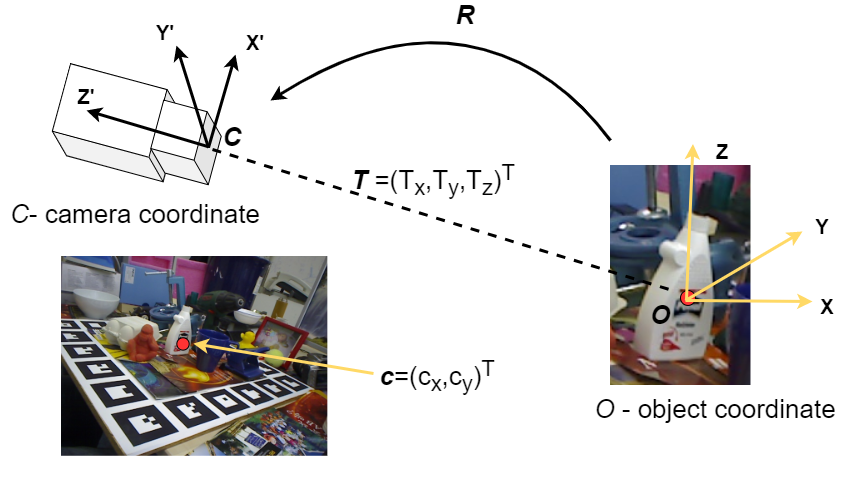}
    \caption{\label{problem_def} The object coordinate system and the camera coordinate system are described. }
  \end{figure} 

The 6D pose essentially means the location of the object in the 3D space i.e; the coordinates of the object, along with the orientation in the 3D space given as the angle of rotation in the 3-axes. Since the coordinates and orientation of anything in the world are defined relative to other object, here we define it relative to the camera coordinates that captures the colour image as well as the depth sensor as shown in Fig. \ref{problem_def} (considering both to be on the same hardware). \par
6D pose estimation then means the prediction of the correct magnitude of translation \textbf{T} as well as the rotation \textbf{R} with the 3-axes concerning the point of reference (in this case the camera sensors ).\par
All the prediction is done using the RGB-D data, RGB means Red, Green, and Blue the 3 primary colours which are used to store the real world image digitally on the computers and the depth information is the distance map from the camera to any location corresponding to that pixel. So, the predictions are done via examining the images of the objects along with the depth information about the same image. \par

So to summarize the problem statement,\textbf{ the 6D estimation of the location of the object along with the orientation (6D means 6 Degree of freedom i.e; the translation on 3-axes as well as the rotation about the 3-axes) in the 3D world of an object whose intrinsic properties and appearance are known to us beforehand using the colour information captured by the camera sensor along with the depth information captured by the Depth sensor and at last its type like whether the object is symmetric or non-symmetric.}

\begin{figure*}[!ht]
    \center
    \includegraphics[width=\textwidth]{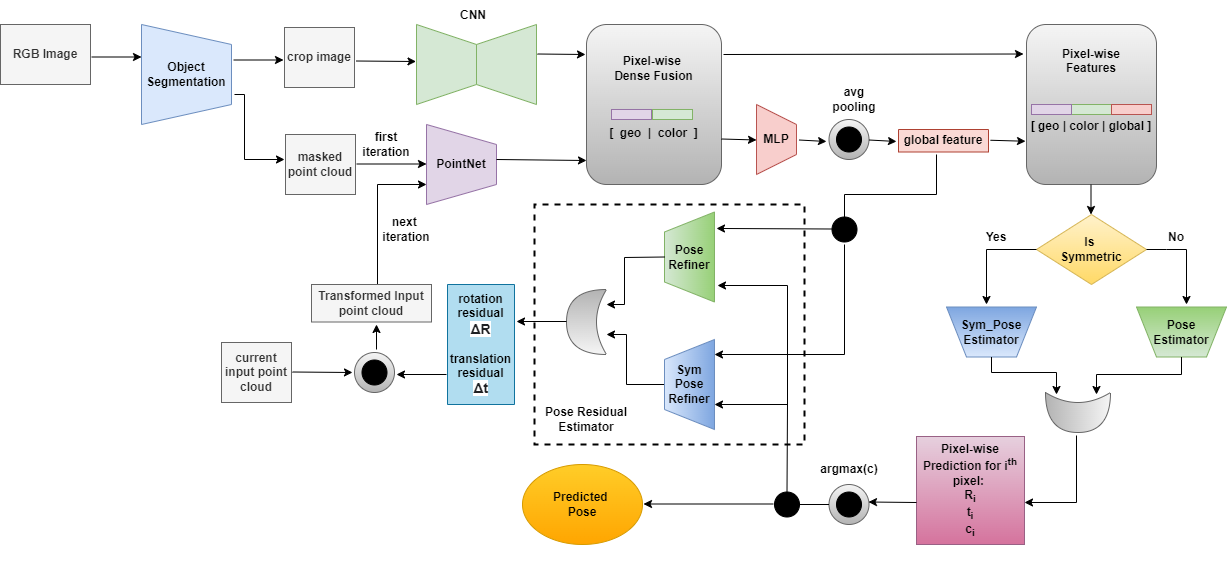}
    \caption{\label{schema} Architectural overview of proposed method}
  \end{figure*}
  
   \section{\textbf{Enhancement}} 
   
   Our work proved an enhanced mechanism over DenseFusion \cite{1} that uses context information that improves the accuracy on the LineMOD dataset. The modified hybrid architecture used has yielded better accuracy. This section explains the logical motivation and the modification in detail.
   \begin{enumerate}[label= (\alph*)]
       \item \textbf{Motivation:} So, the motivation for the current hybrid architecture comes after analysing the intrinsic properties of the types of objects present in the dataset. There are two types of objects present in the dataset, they are symmetric and non-symmetric. There are also two types of loss functions defined for each i.e; \textbf{ADD} for non-symmetric and \textbf{ADD-S} for symmetric objects. Although both types of distances can be taken care of in the same model but they tend to hinder the full potential of the network to learn them. In a way we can infer that the prediction for symmetric objects is a bit less stricter compared to non-symmetric objects. As for symmetric objects, same pose also fits with different orientations if we change the object about the axis of symmetry. This laid the basis for our model.\par
       In our work, when we tried assessing and observing the model, we  observed that for the symmetric objects the learning was faster and also reaching the saturation point in lesser epoch iteration and also observed that the prediction for non-symmetric objects were taking greater time to reach saturation or not attaining the proper saturation in the 6D pose estimator phase (i.e; without refinement step). This laid the idea to train two separate models. Since the extraction and fusion of features had apparently no effect by the type of object, these steps are still common, but the estimation and refinement networks are separated for each type of objects.\\

  \item \textbf{Structure:}  A hybrid architecture is used for the pose estimation and refinement steps. Since we discussed that the learning of non symmetric objects were harder than symmetric, that is why a deeper network is used that consists of ResNet18 network. Additional convolution and pooling layers were added making it deeper, keeping the tradition pooling technique used here intact i.e; the average pooling and finally the output layer outputs the quaternion values, the translation values along with the confidence score for each pixel that helps us choose the best pose after each of the pixel has voted the pose according to their features. For the symmetric objects it was evident from the results published in \cite{1} that the original architecture was good, so similar network is used like \cite{1} with fine tweaking to enhance the prediction. And for the Non-symmetric objects, a deeper network is used in an attempt to learn the information more accurately and without interference from the other. \\
  
  Then after the pose estimation step comes the pose refinement step. Since the refinement model is used to refine the output by the estimator model, a CNN network is used. But there arises a decision  whether to use same refinement network for both estimators or to use separate refiner for each. This decision were made again by considering the fact that both type of objects have their respective intrinsic properties as well as different evaluation measures (loss functions) so separate refinement models were required for both Estimator models. So for both estimators separate refiners were used, basic structure of a typical CNN was used here but the refiner for non-symmetric objects contains additional layers to cope up with the new estimator.\\
  \end{enumerate}
\section{\textbf{Methodology}}

Our main focus in this work is on the 6D pose estimation of known objects that are present in the scene of RGB-D image data. Unlike many other approaches, we are focusing on variable lighting, occluded objects, and cluttered scenes. The first thing to make a model is to represent our output. Basic architecture of DenseFusion \cite{1} is used as a starting point then an enhanced architecture is proposed in this work. Since we are predicting the poses of known objects present in the scene in the image, we need something that can express the orientation and the coordinate in the 3D space referenced about a fixed point. Since our data is captured using Depth and image sensor then we can safely assume that both the sensors are practically at the same point (although the distance between them is not zero but compared to the distance of objects from the camera it could be neglected), so the pose of an object is defined with respect to the coordinates and frame of the camera.\par
The pose of any object needs to be defined before we can prepare to predict it. So for any object, we know its basic structure hence any pose is the combination of rotation in the 3-axes as well as the translation in the 3D space. So, mathematically it can be denoted as p $\in$ SE(3), where p = $[R|t]$ and R $\in$ SO(3) and t $\in \mathbb{R}^3$. \par
As we have discussed up until now that the estimation of 6D poses in the occluded area and poorly or overly exposed to lighting is possible with high efficiency only by combining the depth and color information intelligently. But since both types of data are different, just combining them is not enough instead we need some method that can extract the essence from each type of data and combine them to be meaningful also need the individual essential information contained within them.\\
These challenges were acknowledged and have been pretty accurately taken care in DenseFusion \cite{1}, so our model is based on similar architecture to use the essence of data but in the later stage of the architecture a hybrid model is used. The challenges mentioned above are tackled by :\\
\begin{itemize}
    \item  By a heterogeneous architecture that takes in the colour and depth information separately keeping their individual essential information intact in both the spaces (discussed in section Dense Feature Extraction) and
    \item By intelligently combining their intrinsic mapping between the data sources, it is done by pixel wise dense fusion which utilises the intrinsic camera parameters to map each 3D point cloud to colour pixel.
\end{itemize}

\par
Then after the process described above, we predict the initial poses of objects. There are two separate routes defined for two different kinds of objects that are present in the datasets. First is regular non-symmetric objects i.e; the objects that do not contain any line of symmetry or axis of symmetry at which we can divide the object into two exactly similar sub-parts. These kinds of objects are treated differently than symmetric objects. After the fusion of color and depth information is done by a pixel-wise fusion of both types of data, they are fed to their pose estimator model. For non symmetric objects, the basic ResNet18   with 2 additional layers is used as pose estimator model. For the symmetric objects, only the basic model of ResNet18 is used, as it showed promising performance on symmetric objects in \cite{1}. All these changes were added after observing the nature of both the objects, after training the models for a sufficiently long time we observed that the output accuracy for symmetric objects was getting saturated but for non-symmetric objects, there were still some fluctuations. So separate predictor models were defined and they showed better performance. After the pose prediction, it is further improved by the integrated iterative refinement network. It is a multi layered CNN network followed by 3 fully connected layer and output layer that further refines the pose predicted by the pose estimator model. Each of the types of objects i.e; have their separate refinement network. And this refinement step is integrated with the architecture so no expensive post-processing techniques (\cite{17,16}) are required.

\subsection{\textbf{Overall Architecture}}\label{AA}

The overall architecture shown in Fig. \ref{schema}, can be broken down into two major stages based on the function they perform. The first stage is the semantic segmentation of each known object in the color image. The segmentation provides us with an $(N+1)$ channeled map, where $N = $number of objects in the dataset. This semantic segmentation map is used to generate a bounding box which in turn is used to extract cropped patch of the object. Then this patch along with the masked cloud or masked depth pixel is forwarded to second stage.\\
The second stage is the estimator or predictor stage where actual prediction takes place. This stage further comprises of various steps / components. These components are :\\
\begin{itemize}
\renewcommand\labelitemi{--}
    \item  A CNN based network that works on the images crops generated by semantic segmentation stage and maps the colour information into colour feature embedding.
    \item A PointNet \cite{36} like network that takes in the masked (according to image crop) 3D point cloud as input and gives us the geometric feature embedding.
    \item After the features are embedded from both the spaces i.e; colour and point cloud then we need to combine them to create global features, it is done by a fusion network that performs pixel wise combination and prediction based on self-supervised confidence scoring \cite{7}. But here unlike the Dense Fusion there are separate networks for symmetric and non-symmetric object types that has helped obtain better accuracy.
    \item Then this pose prediction is fed to iterative refinement network in a curriculum learning manner. Here also there are separate iterative refinement networks for both types of objects to support their respective estimators such that each estimator network and iterative network form a separate pair for each type of object.
\end{itemize}
\par
The detailed explanation about each stage and component is given below: 

\subsubsection{ \textbf{ Semantic Segmentation}} 
To predict any object's pose we first need to find out that particular object from the scene. Segmentation helps us to classify a scene into various segments. Semantic segmentation means recognizing and labeling each known object in the scene. Since it is the prerequisite for our stage II and already many efficient models exist for segmentation, a preexisting vanilla segmentation model is used. Vanilla segmentation is an encoder-decoder-based architecture that first takes the color image as input then encodes the information into smaller dimension features then decodes them into N+1 channeled segmentation map where each channel is a binary mask and true pixels indicate the presence of that particular object. One extra channel is to denote the background or no object. Since in this work, the focus is on pose estimation rather than segmentation we use an existing architecture \cite{16}. \\

\subsubsection{\textbf{ Feature Extraction and Pose Estimation}} 
\begin{enumerate}[label=(\alph*)]

    \item  \textbf{Dense Feature Extraction:} Since we are using both image data and depth data, we need to extract meaningful information out of it. Some of the previous methods used depth information as an additional channel and then used CNN based architecture to directly predict poses, but the main shortcoming about this approach is that we are considering both types of data i.e; image and depth to be the same hence neglecting their respective implicit structure even though they lie in different spaces. In DenseFusion \cite{1} this problem is recognized and critical architecture was used which is being reused in our work as well.\\ 
    First, the 3D data is converted into a 3D point cloud by using intrinsic camera properties (a concept from image processing) then PointNet \cite{36} like structure is used. The PointNet was able to perform well in this segment as they used the symmetric max pooling function to get permutation in-variance in unordered point sets. Similar architecture is used here just the symmetric function is replaced by average pooling in place of max pooling which is commonly used.   \\
    Secondly, the colour image data needs to be embedded into features, it is done by using a CNN based encoder decoder network that converts H X W X 3 space into H X W X $d_{rgb}$ space which means each pixel will now get $d_{rgb}$ dimension feature vector. \\

    \item \textbf{Dense Feature Fusion:} Now that we have obtained separate feature embedding, we need to find out a way to combine them to produce effective features. Our discussion up until now have made it clear that many have tried to use the RGB-D data for pose estimation but what is required is actually how we fuse both types of data since they lie in different spaces. The work in DenseFusion \cite{1} proves effective in this particular segment. Instead of treating both RGB and Depth information similar and blindly fusing them, which in turn will result in degraded performance, the proposed a novel pixel-wise fusion network for this purpose.\\
    \\Due to segmentation errors, occlusion, and variable lighting, directly associating color info with corresponding depth info on the same pixel will not retain the 3D behavior. So, projection (image processing concept) is used, which first projects the color information of a pixel in the segment into 3D space by using intrinsic camera parameters then at each pixel associates it with a geometric feature. So now we have got a pair of colors and geometric features. Now that we have got hold of each of the data sources' intrinsic properties, we need a way to extract information that is the property of combined data. This is done by using an MLP network with a symmetric function (in this case average pooling ) and then concatenated to our original per pixel features. So now we have for each pixel color features, geometric features, and global features. This whole process of forming per pixel features is named as Dense Fusion of features in DenseFusion.\\
    Now after obtaining the features, the actual estimation is remaining, the prediction is based on the work of Xu et al. \cite{7} in which for each pixel feature there is a prediction of pose along with a confidence score. Then we predict the final pose in this that has the highest confidence score among all the per-pixel prediction.\\ 
    
    \item \textbf{6D Pose Estimation: } Our work puts emphasis on this component to obtain better accuracy. While the DenseFusion \cite{1} used a single ResNet18 based network for both types of data, we modified its design after observing the difference between symmetric and non-symmetric objects. We can infer that for a symmetric object there can be multiple orientations that are accurate for a particular pose. About the axis of symmetry if we rotate the object then the pose doesn't show any observable change while there is no such property present in non-symmetric, so we can say that in a way predicting the pose of a symmetric object is less strict than non-symmetric. So taking this fundamental difference in mind we propose separate networks for each type of object. More details are explained in the later section.
    
    For the model we first need to define a learning rule and learning rule is always based on the loss function. A loss function is the mathematical expression that calculates how much difference there is in predicted and desired output. The loss function used here is defined as: 


 \begin{equation} \label{eqn:loss1}
  L_{i}^{p} = \frac{1}{M} \sum_{j} || (Rx_{j} + t) - (R_{i}^{\string^}x_{j} + t^{\string^}_{i}) ||
    \end{equation}

   where $x_j$ denotes $j^{th}$ point out of the randomly selected M points from object, p = $[R|t]$ is the desired or actual pose while $p^{\string^}_i = [R^{\string^}_i|t^{\string^}_i]$ is the prediction from the $i^{th}$ pixel features.\\
   But as discussed a moment ago same pose can be applicable to even infinity orientation of symmetric objects so the loss for symmetric function needs to be defined as current one will lead to ambiguity. Therefore different loss function is required for symmetric objects, the one used here is defined as:
 
    \begin{equation} \label{eqn:loss2}
  L_{i}^{p} = \frac{1}{M} \sum_{j} \min_{0<k<M} || (Rx_{j} + t) - (R_{i}^{\string^}x_{k} + t^{\string^}_{i}) ||
    \end{equation}
   
   
 So, now that individual object's loss is defined then we want to define overall loss given by:

    \begin{equation} \label{eqn:loss_overall}
  L = \frac{1}{N} \sum_{i} ( L_{i}^{p}c_{i} - w\log({c_{i}}) )
    \end{equation}

where N is the number of randomly selected pixels feature from P elements of the particular segment and w is regularization term. Using this loss we can balance the highly confident prediction and low confidence prediction along with the loss. With the regularization term the low confidence prediction will get low loss but will incur higher penalty too. After this we choose the pose with the highest confidence score.\\

   \item \textbf{Iterative Refinement:} Many refinement approaches have used post-processing techniques like ICP \cite{18} used in \cite{43,32,16} which showed promising results in pose estimation but due to their costly post-processing nature, they were not fit for real-time applications. The solution was proposed in DenseFusion which is used in our work. DenseFusion uses an integrated CNN based iterative refinement module that can further improve the pose estimation by our previous component. This refinement network's work is not to predict something new but to refine the output by the main network. It takes in as input the pose predicted in the previous iteration then this, along with the global feature from the Feature fusion stage, is used to calculate the pose residual which is then used to convert the input point clouds into the previously predicted pose as an estimate of canonical form. But initially, the training of the refinement network would yield no good results as there will be too much noise present in the initial stage, so it will start after the prediction attains certain accuracy.
   
  \end{enumerate}

   \section{\textbf{Experiments and Performance Analysis}}

\begin{table*}[ht]

\caption{Comparison on LineMOD \cite{20} dataset for 6D pose estimation. Symmetric objects are in bold name.}
\label{lineMOD}
\centering
\scalebox{1}{
\begin{tabular}{|c|c|c|c|c|c|c|c|}
\hline
 & \textbf{RGB} & \multicolumn{6}{|c|}{RGB-D} \\
 \hline
& \textbf{PoseCNN \cite{16} }& \textbf{Implicit} & \textbf{SSD-6D} & \textbf{PointFusion} & \textbf{ DenseFusion} \cite{1} & \textbf{OUR} & \textbf{OUR} \\
& \textbf{+ DeepIM \cite{40}}& \textbf{ \cite{32} + ICP} & \textbf{\cite{25} +ICP} & \textbf{\cite{7}} & \textbf{  (2 iterations)} & \textbf{(2 iterations)} & \textbf{(10 iterations)} \\
\hline
ape         & 77.0 & 20.6 & 65  & 70.4 & 92.3 & 99.05   & 99.05\\
bench vi.   & 97.5 & 64.3 & 80  & 80.7 & 93.2 & 99.03   & 99.03  \\
camera      & 93.5 & 63.2 & 78  & 60.8 & 94.4 & 97.06   & 98.04  \\
can         & 96.5 & 76.1 & 86  & 61.1 & 93.1 & 96.04   & 98.02 \\
cat         & 82.1 & 72.0 & 70  & 79.1 & 96.5 & 96.99   & 96.99  \\
driller     & 95.0 & 41.6 & 73  & 47.3 & 87.0 & 96.99   & 97.99  \\
duck        & 77.7 & 32.4 & 66  & 63.0 & 92.3 & 93.40   & 94.34 \\
\textbf{eggbox}      & 97.1 & 98.6 & 100 & 99.9 & 99.8 & 94.34   & 95.28 \\
\textbf{glue}        & 99.4 & 96.4 & 100 & 99.3 & 100.0 & 90.29  & 95.14  \\
hole p.     & 52.8 & 49.9 & 49  & 71.8 & 92.1 & 98.09   & 98.09  \\
iron        & 98.3 & 63.1 & 78  & 83.2 & 97.0 & 97.94   & 97.94  \\
lamp        & 97.5 & 91.7 & 73  & 62.3 & 95.3 & 98.08   & 99.99  \\
phone       & 87.7 & 71.0 & 79  & 78.8 & 92.8 & 97.11   & 98.07  \\
\hline
MEAN        & 88.6 & 64.7 & 79  & 73.7 & 94.3 & 96.48   & 97.52 \\
\hline
\end{tabular}
}
\end{table*}

 \begin{figure}[ht]
\centerline{\scalebox{0.19}{\includegraphics{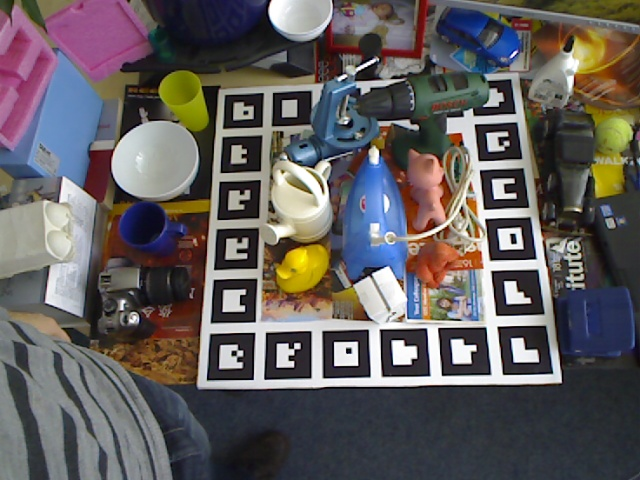} }
\scalebox{0.19}{\includegraphics{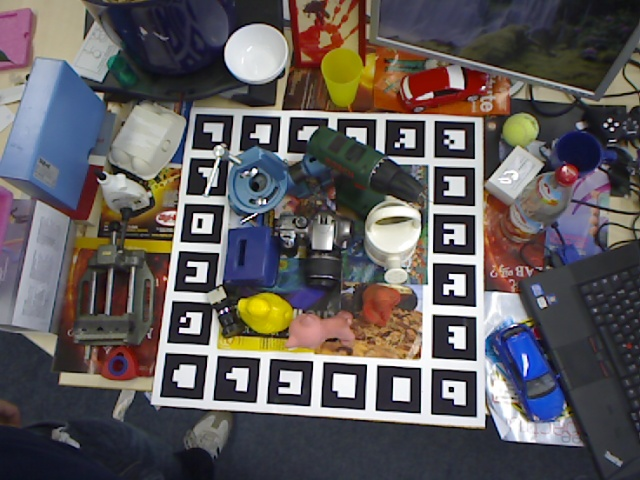}}}
\caption{RGB Samples from LineMOD dataset \cite{20}.}
\label{rgbfig}
\end{figure}

   \begin{figure}[ht]
\centerline{\scalebox{0.38}{\includegraphics{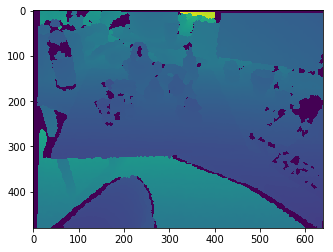} }
\scalebox{0.38}{\includegraphics{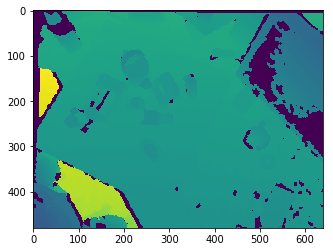}}}
\caption{Depth Samples from LineMOD dataset \cite{20}.}
\label{depthfig}
\end{figure}
   This section is dedicated for the discussion of the experiments conducted, the performance analysis of the model, comparison with preexisting work, the datasets specifications, etc. For the evaluation of the model we need dataset on which the training and testing is to be done, the measurement unit which can express the performance into real numbers, and the comparisons with existing models to compare the accuracy as a benchmark. So all these things are discussed in details below.
   
    \subsection{\textbf{Datasets}}
   
   One of the most popular datasets used for the evaluation of pose estimation task, namely LineMOD \cite{20} dataset. Each of which comprises images videos (which is simply the collection of images) containing some known objects in each image.\\
   \subsubsection{\textbf{LineMOD dataset}} The LineMOD dataset from Hinterstoisser et al. \cite{20} comprises 13 low textured objects spanning 13 videos. Many classical, as well as the modern learning-based, \cite{40,32,25} has adopted this for training, testing, and evaluation purposes. The training and testing partition used is the same as some of the prior learning works \cite{1,40,25} without appending synthetically generated data. The 3D models of objects are also provided in the dataset. This dataset is considered the benchmark for highly cluttered object's pose estimation. Fig. \ref{rgbfig} and Fig. \ref{depthfig} shows RGB samples and Depth samples respectively from LineMOD dataset.


   
   
   \subsection{\textbf{Metrics}}
   There are two matrices to measure the prediction. ADD and ADD-S, both are a measures of average point wise distance. Both of them operates on the predicted pose $[R^{\string^}|t^{\string^}]$ and the ground truth $[R|t]$. 
   
   \subsubsection{\textbf{ADD}} It is the Average distance of Model Points, which means the distance between the predicted location and the actual location of the model points selected randomly. This metric is applicable only for non-symmetric objects as the symmetric object will have ambiguous orientation pertaining to the same pose due to symmetric property. Its mathematical representation is given by  (\ref{eqn:loss1}).
   
   
   
   \subsubsection{\textbf{ADD-S}} ADD-S is similar to ADD only difference being that the distance between predicted and actual is calculated using the closest point only. This makes it fit and non-ambiguous for symmetric objects as well. The ADD-S below 2cm is considered as correctly predicted as it is considered as the threshold for robot manipulation tasks. Mathematically, it is denoted by  (\ref{eqn:loss2}).
   
   
   
   For the LineMOD dataset both metrics are used, ADD for non symmetric and ADD-S for symmetric objects.\\

 \subsection{\textbf{Performance analysis on LineMOD Dataset}  }

\begin{figure}[ht]
\centerline{\scalebox{0.4}{\includegraphics{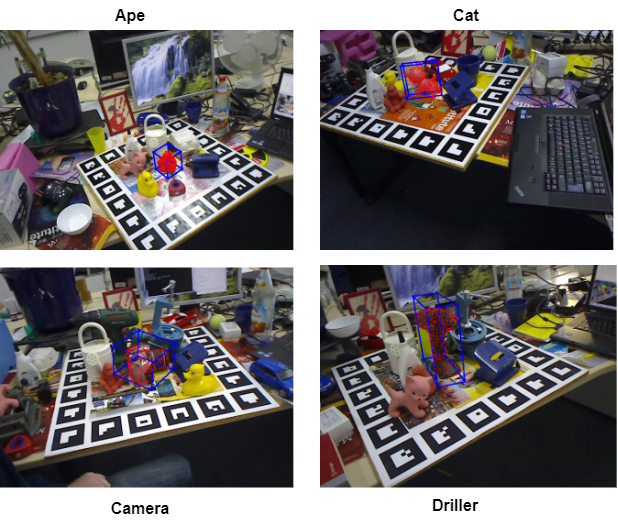} }}
\caption{Result Visualization for LineMOD dataset.}
\label{resultfig}
\end{figure}

\subsubsection{\textbf{Accuracy}}
In  this  section  we  compare  the  performance  of  our  model with that of some of the most popular methods as show in TABLE \ref{lineMOD}. On the LineMOD dataset the DenseFusion’s   accuracy were 86.2\% without refinement and 94.3\% with 2 iterations of refinement which got saturated after 2 iterations, whereas  in  our experiment  we  obtained  \textbf{96.4\%}  accuracy  with  2  refinement steps  while  \textbf{97.5\%}  with  10  iteration  of  refinement,  which  is \textbf{2.2\%} more with 2 refinement iteration and \textbf{3.2\%} more with 10 iteration. In Fig. \ref{resultfig}, we have visualized the estimated 6D pose for some of the objects of the LineMOD dataset.


\subsubsection{\textbf{Time}}
Along with accuracy, time taken to process one frame is also important if we wish to use the model in real time application. So, we also compared the time taken for the model to process and output for one single frame. PoseCNN + ICP \cite{16} took around 10.6 seconds for one frame, DenseFusion \cite{1} took around 0.06 seconds and our model took around 0.065 seconds for the same which is approximately the same as DenseFusion.

\begin{table}[ht]
\caption{Comparison of time taken for each frame (in seconds).}
\label{time}
\centering
\scalebox{1}{
\begin{tabular}{|c|c|c|c|}
\hline
\textbf{PoseCNN + ICP \cite{16}} & \textbf{DenseFusion \cite{1}} & \textbf{Our}\\
\hline
10.6 & 0.06 & 0.065\\
\hline
\end{tabular}
}
\end{table}
So according to the TABLE \ref{time}, our model takes around 0.065 sec which when converted into video frames per sec gives around 13-15 fps which is pretty sufficient for real time applications.

\section*{\textbf{Conclusion}}
In this work,  a hybrid context-aware architecture is introduced that performs the pose estimation of objects in cluttered areas efficiently by using the inherent difference in properties of two types of objects present in the data. It is done by treating symmetric and non-symmetric objects separately. Our method has achieved an accuracy of 97.52\% on the LineMOD dataset, which makes it 3.2\% accurate than  DenseFusion. Our model also keeps the inference time very low to make it an efficient choice for real-time applications.

\section*{\textbf{Acknowledgements}}
The present research is partially funded by the I-Hub foundation for Cobotics (Technology Innovation Hub of IIT-Delhi setup  by the Department of Science and Technology, Govt. of India).

\section*{\textbf{Conflict of interest}}
The authors declare that they have no conflict of interest.



\printbibliography

\end{document}